\documentclass[11pt]{article}
\usepackage{acl2015}
\usepackage{times}
\usepackage{url}
\usepackage{tikz}
\usetikzlibrary{positioning}
\usepackage{dsfont}
\usepackage{latexsym}
\usepackage{graphicx}
\graphicspath{ {pictures/} }
\DeclareGraphicsExtensions{.pdf,.png}
\usepackage{hyperref}
\usepackage{multirow}
\usepackage{amsmath, amssymb, amsfonts}
\usepackage[font={small,it}]{caption}

\title{Class Vectors: Embedding representation of Document Classes}

\author{Devendra Singh Sachan \\ Enixta Innovations \\ Hyderabad, India\\
{\tt devendras@enixta.com} \\\And
Shailesh Kumar \\ Enixta Innovations \\ Hyderabad, India \\
{\tt shkumar@enixta.com} \\}

\begin{document}
\maketitle

\begin{abstract}
Distributed representations of words and paragraphs as semantic embeddings in high dimensional data are used across a number of Natural Language Understanding tasks such as retrieval, translation, and classification. In this work, we propose "Class Vectors" - a framework for learning a vector per class in the same embedding space as the word and paragraph embeddings. Similarity between these class vectors and word vectors are used as features to classify a document to a class.

In experiment on several sentiment analysis tasks such as Yelp reviews and Amazon electronic product reviews, class vectors have shown better or comparable results in classification while learning very meaningful class embeddings.
\end{abstract}

\section{Introduction}
Text classification is one of the important tasks in natural language processing. In text classification tasks, the objective is to categorize documents into one or more predefined classes. This finds application in opinion mining and sentiment analysis (e.g. detecting the polarity of reviews, comments or tweets etc.) \cite{pang_opinion_2008}, topic categorization ( e.g. aspect classification of web-pages and news articles such as sports, technical etc.) and legal document discovery etc. \\

In text analysis, supervised machine learning algorithms such as Naive Bayes (NB) \cite{McCallum98acomparison}, Logistic Regression (LR) and Support Vector Machine (SVM) \cite{Joachims:1998:TCS:645326.649721} are used in text classification tasks. The bag of words \cite{harris54} approach is commonly used for feature extraction and the features can be either binary presence of terms or term frequency or weighted term frequency. It suffers from data sparsity problem when the size of training data is small but it works remarkably well when size of training data is not an issue and its results are comparable with more complex algorithms \cite{sidaw12simple}. \\

Using the co-occurring words information, we can learn distributed representation of words and phrases \cite{morin_hierarchical_2005} in which each term is represented by a dense vector in embedding space. In the skip-gram model \cite{mikolov_distributed_2013}, the objective is to maximize the prediction probability of adjacent surrounding words given current word while global-vectors model \cite{pennington-socher-manning:2014:EMNLP2014} minimizes the difference between dot product of word vectors and the logarithm of words co-occurrence probability. \\

One remarkable property of these vectors is that they learn the semantic relationships between words i.e. in the embedding space, semantically similar words will have higher cosine similarity. For example, the word "\textit{gpu}" will be more similar to "\textit{processor}" than to "\textit{camera}". To use these word vectors in classification tasks, Le et al. \shortcite{le_distributed_2014} proposed the Paragraph Vectors approach, in which they learn the vectors representation for documents by stochastic gradient descent and the gradient is computed by backpropagation of the error from the word vectors. The document vectors and the word vectors are learned jointly. Kim \shortcite{DBLP:journals/corr/Kim14f} demonstrated the application of Convolutional Neural Networks in sentence classification tasks using the pre-trained word embeddings. \\

Taking inspiration from the paragraph vectors approach, we propose class vectors method in which we learn a vector representation for each class. These class vectors are semantically similar to vectors of those words which characterizes the class and also give competitive results in document classification tasks.

\section{Model}
We use skip-gram model \cite{mikolov_distributed_2013} to learn these vectors. In the skip-gram approach, we learn the parameters of model to maximize the prediction probability of the cooccurence of words. Let the words in the corpus be represented as $w_1$, $w_2$, $w_3$, .., $w_n$. The objective function is defined as,
\begin{equation}
L = \sum_{i=1}^{N_s}  \sum_{c \epsilon [-w,w], c\neq 0} \log p(w_{i+c}/w_i)     \label{eq:1}
\end{equation}
where $N_s$ is the number of words in the sentence(corpus) and $L$ denotes the likelihood of the observed data. $w_t$ denotes the current word, while $w_{t+c}$ is the context word within a window of size $w$. The prediction probability $p(w_{i+c}/w_i)$ is calculated using the softmax classifier as below,
\begin{equation}
p(w_{i+c}/w_i) = \frac{\exp\left({v_{w_i}^\intercal v'_{w_{i+c}}}\right)}
{\sum_{w=1}^{T}\exp\left(v_{w_i}^\intercal {v'_{w}}\right)}
\end{equation}

$T$ is number of unique words selected from corpus in the dictionary, $v_{w_i}$ is the vectors representation of the current word from inner layer of neural network while ${v'_{w}}$ is the vector representation of the context word from the outer layer of the neural network. In practice, since the size of dictionary can be quite large, the cost of computing the denominator in the above equation can be very expensive and thus gradient update step becomes impractical. \\ 

Morin et al. \shortcite{morin_hierarchical_2005} proposed Hierarchical Softmax to speed up the training. They construct a binary Huffman tree to compute the probability distribution which gives logarithmic speedup $\log_2(T)$. Mikolov et al. \shortcite{mikolov_distributed_2013} proposed negative sampling which approximates $\log{p(w_{i+c}/w_i)}$ as,
\begin{equation}
   \log \sigma({v_{w_i}^\intercal v'_{w_{i+c}}} ) + \sum_{j=1}^k\mathbb E_{w_j\sim
    P_n(w)}\left(\log \sigma(-{v_{w_i}}^\intercal v'_{w_j})\right)
\end{equation}

$\sigma(x)$ is the sigmoid function, the word ${w_j}$ is sampled from probability distribution over words ${P_n(w)}$. The word vectors are updated by maximizing the likelihood $L$ using stochastic gradient ascent. \\

Our model, shown in Figure 1, learns a vector representation for each of the classes along with word vectors in the same embedding space. We represent each class vector by its id (class\_id). Each class id co-occurs with every sentence and thus with every word in that class. Basically, each class id has a window length of the number of words in that class. We call them as Class Vectors (CV). Following eq\ref{eq:1} new objective function becomes,
\begin{equation}
    \sum_{i=1}^{N_s}  \sum_{c \epsilon [-w,w], c\neq 0} \log p(w_{i+c}/w_i) + 
    \lambda \sum_{j=1}^{N_c} \sum_{i=1}^{N_j} \log p(w_{i}/c_{j})
\end{equation}
$N_c$ is the number of classes, $N_j$ is the number of words in $class_j$, $c_j$ is the class id of the $class_j$. We use skipgram method to learn both the word vectors and class vectors.\\

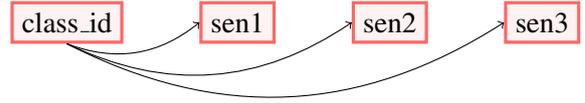
\begin{figure}[h!]
\centering
\begin{tikzpicture}[
roundnode/.style={circle, draw=green!60, fill=green!5, very thick, minimum size=7mm},
squarednode/.style={rectangle, draw=red!60, fill=red!5, very thick, minimum width=5mm, minimum height = 3mm},
myrect/.style={rectangle, draw, inner sep=0pt, fit=#1}
]
\node[squarednode]      (class)                 {class\_id};
\node[squarednode]      (word1)       [right=of class] {sen1};
\node[squarednode]      (word2)       [right=of word1] {sen2};
\node[squarednode]      (word3)       [right=of word2] {sen3};
 
\draw[bend right, ->] (class.south) to node [auto] {} (word1.west);
\draw[bend right, ->] (class.south) to node [auto] {} (word2.west);
\draw[bend right, ->] (class.south) to node [auto] {} (word3.west);
\end{tikzpicture}
\caption{Class Vectors model. While training each class vector is represented by an id. Every word in the sentence of that class co-occurs with its class vector. Class vectors and words vectors are jointly trained using skip-gram approach. }
\end{figure}

\subsection{Class Vector based scoring} \label{Class Vector based scoring}
Converting class vector to word similarity to probabilistic score using softmax function

\begin{equation}
s(w_j/c_i) = \frac{\exp\left({v_{c_i}^\intercal v_{w_j}}\right)}
{\sum_{w=1}^{T}\exp\left(v_{c_i}^\intercal v_{w_j}\right)}
\end{equation}
$v_{c_i}$ and $v_{w_j}$ are the inner un-normalised $ith$ class vector and $jth$ word vector respectively. To predict the class of test data, we use different ways as described below

\begin{itemize}
\item We do summation of probability score for all the words in sentence for each class and predict the class with the maximum score. \textbf{(CV Score)}
\begin{equation}
\underset{i=1,..,C}{\operatorname{arg\,max}} {\sum_{j=1}^{N_s}\log(s(w_j/c_i))}
\end{equation}

\item We take the difference of the probability score of the class vectors and use them as features in the bag of words model followed by Logistic Regression classifier. For example, in the case of sentiment analysis, the two class are \textit{positive} and \textit{negative}. So, the expression becomes,  \textbf{(CV-LR)}
\begin{equation}
f(w) = \log(s(w/c_{pos})) - \log(s(w/c_{neg}))
\end{equation}
$w$ is the vector of the words in vocabulary.

\item We compute the similarity between class vectors and word vectors after normalizing them by their \textit{l2-norm} and using the difference between the similarity score as features in bag of words model. \textbf{(norm CV-LR)}
\begin{equation}
f(w) = {v_{c_{pos}}^\intercal v_{w}} - {v_{c_{neg}}^\intercal v_{w}}
\end{equation}

\end{itemize}

\subsection{Feature Selection}
Important features in the corpus can be selected by information theoretic criteria such as conditional entropy and mutual information. We assume the entropy of the class to be maximum i.e. $H(C) = 1$ irrespective of the number of documents in each class. Realized information of class given a feature $w_i$ is defined as,
\begin{equation}
    I(C;w=w_i) = H(C) - H(C/w=w_i)
\end{equation}
where conditional entropy of class, $H(C/w_i)$ is,
\begin{equation}
    H(C/w=w_i) = -\sum_{c_i}^{N_c}p(c_i/w_i)\log_2 p(c_i/w_i)
\end{equation}

\begin{equation}
    p(c/w_i) = \frac{\exp\left({v_{c_i}^\intercal v_{w_i}}\right)}
                    {\sum_{c_i}^{N_c}\exp\left(v_{c_i}^\intercal v_{w_i}\right)}
\end{equation}
We calculate expected information $I(C;w)$ also called mutual information for each word as,
\begin{equation}
    I(C;w) = H(C) - \sum_{w}p(w)H(C/w)
\end{equation}
$p(w)$ is calculated from the document frequency of word. 
We plot expected information vs realized information to see the important features in the dataset.

\section{Dataset description}

We did experiments on Amazon Electronic Reviews corpus and Yelp Restaurant Reviews. The task is to do sentiment classification among 2 classes ( i.e. each review can belong to either positive class or negative class ) .

\begin{itemize}
\item \textbf{Amazon Electronic Product reviews}
\footnote{\url{http://riejohnson.com/cnn_data.html}} - This dataset is a part of large Amazon reviews dataset McAuley et al.,\shortcite{McALes13b}\footnote{\url{http://snap.stanford.edu/data/web-Amazon.html}}. This dataset \cite{johnson-zhang:2015:NAACL-HLT} contains training set of 392K reviews split into various various sizes and a test set of 25K reviews. We pre-process the data by converting the text to lowercase and removing some punctuation characters.

\item \textbf{Yelp Reviews corpus} 
\footnote{\url{https://www.kaggle.com/c/yelp-recruiting/data}} - This reviews dataset was provided by Yelp as a part of Kaggle competition. Each review contains star rating from 1 to 5. Following the generation of above IMDB Movie Reviews and Amazon Electronic Product Reviews data we considered ratings 1 and 2 as negative class and 4 and 5 as positive class. We separated the files into ratings and do pre-processing of the corpus. \footnote{We use the code available at \url{https://github.com/TaddyLab/deepir/blob/master/code/parseyelp.py}} \cite{Matt-Taddy} 
In this way, we obtain around 193K reviews for training and around 20K reviews for testing.

\end{itemize}

\begin{table}[ht]
\hspace{-.25cm}
{
\begin{tabular}{| c |c | c | c |}
\hline
\textbf{Dataset} & \textbf{Pos Train} & \textbf{Neg Train} & \textbf{Test Set} \\
\hline
Amazon & 196000 & 196000 & 25000 \\
\hline
Yelp & 154506 & 38172 & 19931 \\
\hline
\end{tabular}}
\caption{ Dataset summary. \textbf{Pos Train}: Number of training examples in positive class. \textbf{Neg Train}: Number of training examples in negative class. \textbf{Test Set}: Number of reviews in Test Set}
\end{table}

\section{Experiments}

We do phrase identification in the data by two sequential iterations using the approach as described in Kumar et al. \shortcite{kumar2014phrase}. We select the top important phrases according to their frequency and coherence and annotate the corpus with phrases. To do experiments and train the models, we consider those words whose frequency is greater than 5. We use this common setup for all the experiments. \\

We did experiments with following methods. In the \textbf{bag of words}(bow) approach in which we annotate the corpus with phrases as mentioned earlier. We report the best results among the bag of words in table 2. In the bag of words method, we extract the features by using
\begin{enumerate}
  \item presence/absence of words \textbf{(binary)}
  \item term frequency of the words \textbf{(tf)}
  \item inverse document frequency of words \textbf{(idf)}
  \item product of term frequency and inverse document frequency of words \textbf{(tf-idf)}
\end{enumerate}  
  
We also evaluate some of the recent state of the art methods for text classification on the above datasets
\begin{enumerate}
    \item naive bayes features in bag of words followed by Logistic Regression \textbf{(NB-LR)} \cite{sidaw12simple}
  \item inversion of distributed language representation \textbf{(W2V inversion)} \cite{Matt-Taddy} \footnote{We use the code available at \url{https://github.com/TaddyLab/deepir} which builds on top of gensim toolkit \cite{rehurek_lrec}}
  \item Convolutional Neural Networks for text categorization \textbf{(CNN)} \cite{johnson-zhang:2015:NAACL-HLT}
  \item Paragraph Vectors - Distributed Bag of Words Model \textbf{(PV-DBOW)} \cite{le_distributed_2014}
\end{enumerate}

Class Vector method based scoring and feature extraction. We extend the open-source code \url{https://code.google.com/p/word2vec/} to implement the class vectors approach. We learn the class vectors and word embeddings using these hyperparameter settings (\textit{window=10, negative=5, min\_count=5, sample=1e-3, hs=1, iterations=40, $\lambda$=1}).  For prediction, we experiment with the three approaches as mentioned above. (\ref{Class Vector based scoring}) \\ 

After the features are extracted we train Logistic Regression classifier in scikit-learn \cite{scikit-learn} to compute the results. \footnote{\url{http://scikit-learn.org/stable/modules/generated/sklearn.linear_model.LogisticRegression.html}} Results of our model and other models are listed in table 2. \\

\begin{table}
\centering
\hspace{-.25cm}
{
\begin{tabular}{ | c | c | c | }
\hline
\textbf{Model} & \textbf{Amazon} & \textbf{Yelp} \\
\hline
bow binary & 91.29 & 92.48 \\
\hline
bow tf & 90.49 & 91.45 \\
\hline
bow idf & 92.00 & 93.98 \\
\hline
bow tf-idf & 91.76 & 93.46 \\
\hline
Naive Bayes & 86.25 & 89.77 \\
\hline
NB-LR & 91.49 & 94.68 \\
\hline
W2V inversion & -- & 93.3 \\
\hline
CNN & \textbf{92.86} & -- \\
\hline
PV-DBOW & 90.07 & 92.86 \\
\hline
CV Score & 84.06 & 87.85 \\
\hline
norm CV-LR & 91.58 & \textbf{94.91} \\
\hline
CV-LR & 91.70 & \textbf{94.83} \\
\hline
\end{tabular}}
\centering
\captionsetup{justification=centering}
\caption{ Comparison of accuracy scores for different algorithms }
\end{table}

\begin{figure*}
    \centering
    \includegraphics[width=1\textwidth]{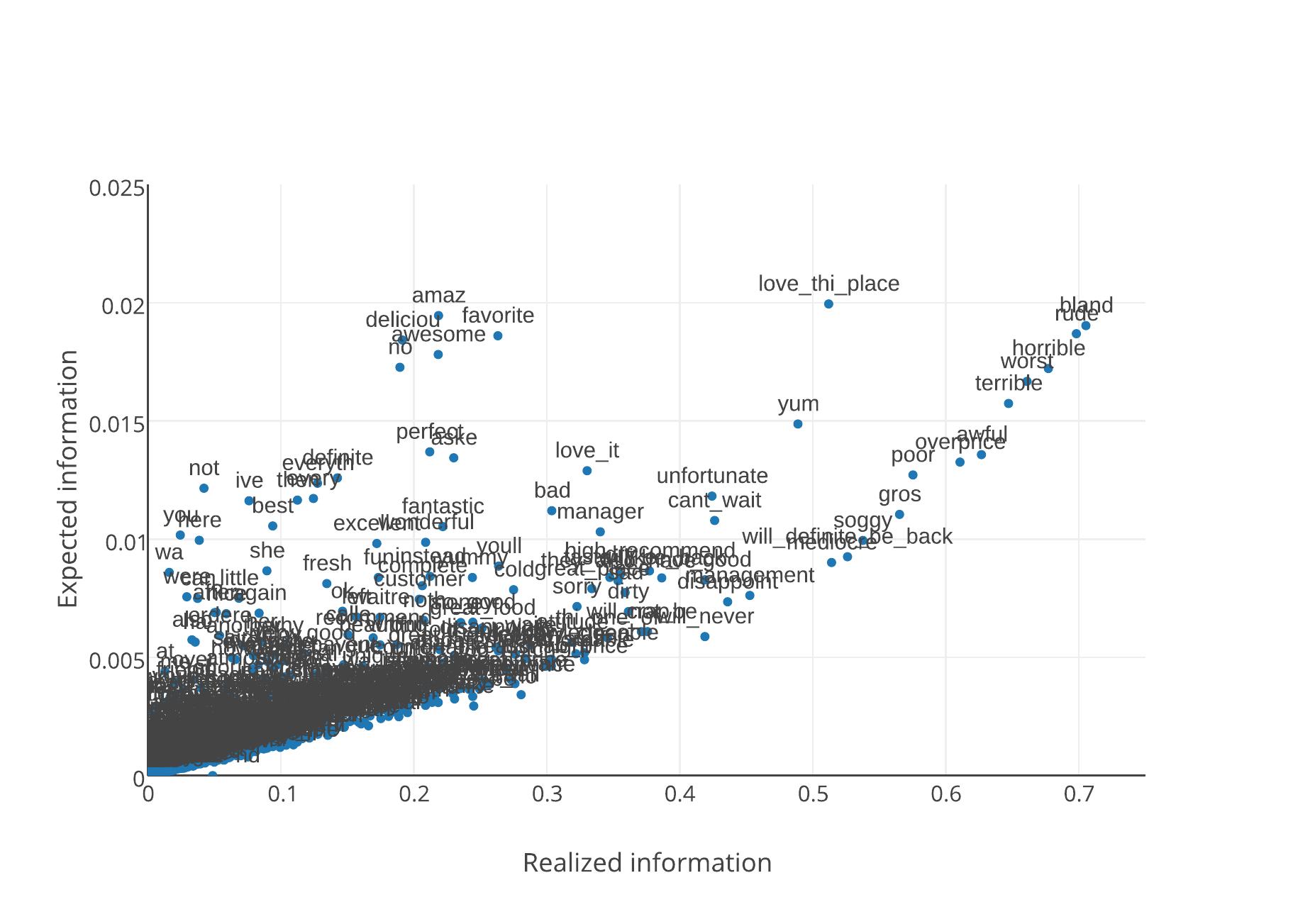}
    \vskip -.25cm
    \caption{Expected information vs Realized information using normalized vectors for $1500$ most frequent words in Yelp Reviews Corpus}
    \label{fig:my_label}
\end{figure*}

\section{Results and Discussion}
\begin{enumerate}
\item We found that annotating the corpus by phrases is important to give better results. For example, the accuracy of PV-DBOW method on Yelp Reviews increased from 89.67\% (without phrases) to 92.86\% (with phrases)  which is more than 3\% increase in accuracy .
\item Class vectors have high cosine similarity with words which discriminate between classes. For example, when trained on Yelp reviews, positive class vector was similar to words like "\textit{very\_very\_good}", "\textit{fantastic}" while negative class vector was similar to words like "\textit{awful}", "\textit{terrible}" etc. More results can be seen in Table 3 and Table 4.
\item In Figure 2, we see that class informative words have greater values of both expected information and realized information. One advantage of class vectors based feature selection method over document frequency based method is that low frequency words can have high mutual information value.
\item On Yelp reviews dataset, we find that the class vectors based approach (CV-LR and norm CV-LR) performs much better than normalized term frequency (tf), tf-idf weighted bag of words, paragraph vectors and W2V inversion and it achieves competitive results in sentiment classification.
\item On Amazon reviews dataset, bow idf performs surprisingly well and outperforms all other methods except CNN based approach. 
\item Shuffling the corpus is important to learn high quality class vectors. When learning the class vectors using only the data of that class, we find that class vectors lose their discriminating power. So, it is important to jointly learn the model using full dataset.  
 
\end{enumerate}

\begin{table}
\centering
\hspace{-.25cm}
{
\begin{tabular}{ | c | c | }
\hline
\multicolumn{2}{|c|}{\textbf{Amazon Electronic Product Reviews}} \\
\hline
\multicolumn{2}{|c|}{\textbf{Top Similar Words to}} \\
\hline
\textbf{Pos class vector} & \textbf{Neg class vector} \\
\hline
very\_pleased &  unfortunately\\
product\_works\_great & very\_disappointed \\
awesome & piece\_of\_crap \\
more\_than\_i\_expected & piece\_of\_garbage \\
very\_satisfied & hunk\_of\_junk \\
great\_buy & awful\_service \\
so\_good & even\_worse \\
great\_product & sadly \\
very\_happy & worthless \\
am\_very\_pleased & terrible \\
a\_great\_value & useless \\
it\_works\_great & never\_worked \\
works\_like\_a\_charm & horrible \\
great\_purchase & terrible\_product \\
fantastic & wasted\_my\_money \\
\hline
\end{tabular}}

\centering
\captionsetup{justification=centering}
\caption{ Top 15 similar words to the positive class vector and negative class vector. }
\end{table}

\begin{table}
\centering
\hspace{-.25cm}
{
\begin{tabular}{ | c | c | }
\hline
\multicolumn{2}{|c|}{\textbf{Yelp Restaurant Reviews}} \\
\hline
\multicolumn{2}{|c|}{\textbf{Top Similar Words to}} \\
\hline
\textbf{Pos class vector} & \textbf{Neg class vector} \\
\hline
very\_very\_good &  awful\\
fantastic & terrible \\
awesome & horrible \\
amaz & fine\_but \\
very\_yummy & food\_wa\_cold \\
great\_too & awful\_service \\
excellent & horrib \\
real\_good & not\_very\_good \\
spot\_on & pathetic \\
great & tastele \\
food\_wa\_fantastic & mediocre\_at\_best \\
very\_good\_too & unacceptable \\
love\_thi\_place & disgust \\
food\_wa\_awesome & food\_wa\_bland \\
very\_good & crappy\_service \\
\hline
\end{tabular}}

\centering
\captionsetup{justification=centering}
\caption{ Top 15 similar words to the positive class vector and negative class vector. }
\end{table}

\section{Conclusion and Future Work}
We learned the class vectors and used its similarity with words in vocabulary as features effectively in text categorization tasks. \newline

There is a lot of scope for further work and research such as using pre-trained word vectors to compute the class vectors. This will help in cases when training data is small. In order to use more than 1-gram as features we need approaches to compute the embeddings of n-grams from the composition of its uni-grams. Recursive Neural Networks of Socher et al \shortcite{socher_recursive_2013} can be applied in these case. We can also work on generative models of class based on word embeddings and its application in text clustering and text classification.

\section{Acknowledgements}
We thank the anonymous reviewers for great discussion and feedback.

\bibliographystyle{acl}
\bibliography{main}

\begin{thebibliography}{}

\bibitem[\protect\citename{Harris}1954]{harris54}
Zellig Harris.
\newblock 1954.
\newblock Distributional structure.
\newblock {\em Word}, 10(23):146--162.

\bibitem[\protect\citename{Joachims}1998]{Joachims:1998:TCS:645326.649721}
Thorsten Joachims.
\newblock 1998.
\newblock Text categorization with suport vector machines: Learning with many
  relevant features.
\newblock In {\em Proceedings of the 10th European Conference on Machine
  Learning}, ECML '98, pages 137--142, London, UK, UK. Springer-Verlag.

\bibitem[\protect\citename{Johnson and
  Zhang}2015]{johnson-zhang:2015:NAACL-HLT}
Rie Johnson and Tong Zhang.
\newblock 2015.
\newblock Effective use of word order for text categorization with
  convolutional neural networks.
\newblock In {\em Proceedings of the 2015 Conference of the North American
  Chapter of the Association for Computational Linguistics: Human Language
  Technologies}, pages 103--112, Denver, Colorado, May--June. Association for
  Computational Linguistics.

\bibitem[\protect\citename{Kim}2014]{DBLP:journals/corr/Kim14f}
Yoon Kim.
\newblock 2014.
\newblock Convolutional neural networks for sentence classification.
\newblock {\em CoRR}, abs/1408.5882.

\bibitem[\protect\citename{Kumar}2014]{kumar2014phrase}
S.~Kumar.
\newblock 2014.
\newblock Phrase identification in a sequence of words, November~18.
\newblock US Patent 8,892,422.

\bibitem[\protect\citename{Le and Mikolov}2014]{le_distributed_2014}
Quoc~V. Le and Tomas Mikolov.
\newblock 2014.
\newblock Distributed representations of sentences and documents.
\newblock In {\em Proceedings of the 31 st {International} {Conference} on
  {Machine} {Learning}}.

\bibitem[\protect\citename{McAuley and Leskovec}2013]{McALes13b}
J.~J. McAuley and J.~Leskovec.
\newblock 2013.
\newblock Hidden factors and hidden topics: understanding rating dimensions
  with review text.
\newblock In {\em Recommender Systems}.

\bibitem[\protect\citename{McCallum and Nigam}1998]{McCallum98acomparison}
Andrew McCallum and Kamal Nigam.
\newblock 1998.
\newblock A comparison of event models for naive bayes text classification.

\bibitem[\protect\citename{Mikolov \bgroup et al.\egroup
  }2013]{mikolov_distributed_2013}
Tomas Mikolov, Ilya Sutskever, Kai Chen, Greg~S. Corrado, and Jeff Dean.
\newblock 2013.
\newblock Distributed representations of words and phrases and their
  compositionality.
\newblock In {\em Advances in {Neural} {Information} {Processing} {Systems}},
  pages 3111--3119.

\bibitem[\protect\citename{Morin and Bengio}2005]{morin_hierarchical_2005}
Frederic Morin and Yoshua Bengio.
\newblock 2005.
\newblock Hierarchical probabilistic neural network language model.
\newblock In {\em Proceedings of the {International} {Workshop} on {Artificial}
  {Intelligence} and {Statistics}}, pages 246--252.

\bibitem[\protect\citename{Pang and Lee}2008]{pang_opinion_2008}
Bo~Pang and Lillian Lee.
\newblock 2008.
\newblock Opinion {Mining} and {Sentiment} {Analysis}.
\newblock {\em Foundations and Trends in Information Retrieval}, 1-2:1--135.

\bibitem[\protect\citename{Pedregosa \bgroup et al.\egroup }2011]{scikit-learn}
F.~Pedregosa, G.~Varoquaux, A.~Gramfort, V.~Michel, B.~Thirion, O.~Grisel,
  M.~Blondel, P.~Prettenhofer, R.~Weiss, V.~Dubourg, J.~Vanderplas, A.~Passos,
  D.~Cournapeau, M.~Brucher, M.~Perrot, and E.~Duchesnay.
\newblock 2011.
\newblock Scikit-learn: Machine learning in {P}ython.
\newblock {\em Journal of Machine Learning Research}, 12:2825--2830.

\bibitem[\protect\citename{Pennington \bgroup et al.\egroup
  }2014]{pennington-socher-manning:2014:EMNLP2014}
Jeffrey Pennington, Richard Socher, and Christopher Manning.
\newblock 2014.
\newblock Glove: Global vectors for word representation.
\newblock In {\em Proceedings of the 2014 Conference on Empirical Methods in
  Natural Language Processing (EMNLP)}, pages 1532--1543, Doha, Qatar, October.
  Association for Computational Linguistics.

\bibitem[\protect\citename{{\v R}eh{\r u}{\v r}ek and Sojka}2010]{rehurek_lrec}
Radim {\v R}eh{\r u}{\v r}ek and Petr Sojka.
\newblock 2010.
\newblock {Software Framework for Topic Modelling with Large Corpora}.
\newblock In {\em {Proceedings of the LREC 2010 Workshop on New Challenges for
  NLP Frameworks}}, pages 45--50, Valletta, Malta, May. ELRA.
\newblock \url{http://is.muni.cz/publication/884893/en}.

\bibitem[\protect\citename{Socher \bgroup et al.\egroup
  }2013]{socher_recursive_2013}
Richard Socher, Alex Perelygin, Jean~Y. Wu, Jason Chuang, Christopher~D.
  Manning, Andrew~Y. Ng, and Christopher Potts.
\newblock 2013.
\newblock Recursive deep models for semantic compositionality over a sentiment
  treebank.
\newblock In {\em Proceedings of the conference on empirical methods in natural
  language processing ({EMNLP})}, volume 1631, page 1642.

\bibitem[\protect\citename{Taddy}2015]{Matt-Taddy}
Matt Taddy.
\newblock 2015.
\newblock Document classification by inversion of distributed language
  representations.
\newblock In {\em Proceedings of the 53rd Annual Meeting of the Association for
  Computational Linguistics}.

\bibitem[\protect\citename{Wang and Manning}2012]{sidaw12simple}
Sida~I. Wang and Christopher~D. Manning.
\newblock 2012.
\newblock Baselines and bigrams: Simple, good sentiment and topic
  classification.
\newblock In {\em Proceedings of the ACL}, pages 90--94.

\end{thebibliography}

\end{document}